\documentclass[conference]{IEEEtran}
\IEEEoverridecommandlockouts
\usepackage{cite}
\usepackage{amsmath,amssymb,amsfonts}
\usepackage{algorithmic}
\usepackage{graphicx}
\usepackage{textcomp}
\usepackage{xcolor}
\def\BibTeX{{\rm B\kern-.05em{\sc i\kern-.025em b}\kern-.08em
    T\kern-.1667em\lower.7ex\hbox{E}\kern-.125emX}}

\DeclareMathOperator*{\argmax}{arg\,max}

\usepackage{blindtext}
\usepackage{algorithm}
\usepackage{graphbox}

\begin{document}

\title{
An Efficient Approach for Cooperative \\ Multi-Agent Learning Problems
}
\author{
\IEEEauthorblockN{1\textsuperscript{st}Ángel Aso-Mollar}
\IEEEauthorblockA{\textit{Valencian Research Institute for AI (VRAIN)} \\
\textit{Universitat Politècnica de València}\\
Valencia, Spain \\
aaso@vrain.upv.es}
\and
 \IEEEauthorblockN{2\textsuperscript{nd} Eva Onaindia}
\IEEEauthorblockA{\textit{Valencian Research Institute for AI (VRAIN)} \\
\textit{Universitat Politècnica de València}\\
Valencia, Spain \\
onaindia@dsic.upv.es}

}

\maketitle

\begin{abstract}

In this article, we propose a centralized Multi-Agent Learning framework for learning a policy that models the simultaneous behavior of multiple agents that need to coordinate to solve a certain task. Centralized approaches often suffer from the explosion of an action space that is defined by all possible combinations of individual actions, known as joint actions. Our approach addresses the coordination problem via a sequential abstraction, which overcomes the scalability problems typical to centralized methods. It introduces a meta-agent, called \textit{supervisor}, which abstracts joint actions as sequential assignments of actions to each agent. This sequential abstraction not only simplifies the centralized joint action space but also enhances the framework's scalability and efficiency. Our experimental results demonstrate that the proposed approach successfully coordinates agents across a variety of Multi-Agent Learning environments of diverse sizes.

\end{abstract}



\section{Introduction}
\label{introduction}

Cooperative Multi-Agent Systems (MAS) focus on the interaction and coordination of autonomous agents to perform tasks more efficiently than individual agents working in isolation. This involves designing protocols and strategies that allow agents to share information, allocate resources, and synchronize their actions effectively. 

A subclass of problems in cooperative MAS, known as cooperative multi-agent planning, involves multiple planning entities with distributed knowledge or capabilities that attempt to achieve a set of goals \cite{Torre_o_2017}. Another type of problem, where the domain model is inaccessible to the agents and so they cannot reason about it, deals with coordinating the behavior of multiple learning agents that coexist in an environment and work together to solve a task. The field that studies techniques for solving this latter type of problem is called Multi-Agent Learning (MAL), which introduces the challenge of learning to coordinate multiple entities to solve specific tasks like distributed control \cite{stephan2000}, robotic teams \cite{BOWLING2002215}, and stock trading \cite{hsu2001}, among others. 

Distributed and centralized MAL are two different ways of coordinating multiple learning agents in a shared environment. In distributed MAL, agents learn and make decisions independently. This approach improves scalability, resilience, and adaptability, allowing agents to operate with partial information and adapt to local changes. However, due to its decentralized nature, it can struggle with global consistency and optimization. Applications of distributed MAL include online resource allocation \cite{distributedresourcealloc}, distributed learning of a single policy \cite{HEREDIA2019363}, and decentralization with networked agents \cite{zhang2018fully}.

In contrast, centralized MAL involves a single entity collecting and processing information from all agents, and disseminating instructions to ensure a coherent and synchronized strategy. Centralized MAL can benefit from comprehensive data and robust decision-making. However, it can also face bottlenecks and scalability issues. Applications for centralized MAL include centralized training with decentralized execution for the StarCraft multi-agent challenge \cite{zhou2023centralized}, counterfactual multi-agent policy gradients for simulations in autonomous vehicles \cite{counterfactual}, and centralized teaching for combat-like domains \cite{centralizedteaching}.

Both distributed and centralized MAL have their strengths and weaknesses, and whether to choose one over the other depends on factors such as the complexity of the task, the environment, and the need for scalability and robustness. Overall, a centralized approach could outperform a distributed one due to its access to global information, if not for the challenges of scalability and complexity.

The objective of this work is to propose a centralized MAL framework for learning a policy that models the behavior of multiple agents, enabling the resolution of a coordination problem while also addressing scalability issues inherent to centralized MAL. To this end, we propose an approach that transforms the multi-agent problem into a single-agent problem. This transformation allows us to more effectively manage the complexity associated with centralizing the behavior of a large number of agents. Our methodology involves compiling the multi-agent problem into an abstract Markov Decision Process (MDP) \cite{mdpsbasic}. The core idea is to abstract the unknown individual dynamics of the agents into a single high-level MDP that encapsulates the behaviors of all agents. This transformation addresses the scalability issues of centralized approaches, as the new entity focuses solely on assigning and directing actions, significantly reducing the action space.

Traditional centralized MAL approaches are based on combining the independent actions of each agent into joint actions. In contrast, we develop a simplified representation that allows us to train a centralized policy using Reinforcement Learning (RL) \cite{sutton98}. The choice of RL is driven by its model-free nature, which is particularly advantageous when the dynamics of the agents' underlying MDPs are unknown, as in this case. Therefore, we focus our work on the MAL subfield of Multi-Agent Reinforcement Learning (MARL) \cite{surveyMARL08}.

Ultimately, our goal is to alleviate the scalability problems in centralized MARL by reducing the emergent complexity and improving the efficiency of centralized control through the use of a coordinating abstract agent called \textit{supervisor}, whose policy is trained with Deep Reinforcement Learning (DRL) techniques. Our proposal not only alleviates the scalability issues in MARL, but also opens up new possibilities for managing large-scale multi-agent systems, as DRL has been successfully applied in a wide range of complex applications \cite{silver2017mastering, alphago,gu2016deep}.

This paper is structured as follows; section \ref{background} presents a brief background of the techniques used in this work. In section \ref{supervision}, we will introduce and exemplify our approach. Section \ref{implementation} discusses the implementation of our approach and section \ref{experiments} presents various experiments to validate its feasibility and effectiveness. Finally, section \ref{conclusion} concludes the principal ideas of this work.


\section{Background}
\label{background}

The basic concepts on which our approach is based are presented in this section.

\subsection{Reinforcement Learning}

Reinforcement Learning is a computational approach to learning from environmental interaction \cite{sutton98}. The objective of an RL agent is to learn a \textit{policy}, or behavior, maximizing a reward function from the interaction of the agent with the environment along time.

RL scenarios are often modeled as finite Markov Decision Processes (MDP) \cite{mdpsbasic}. An MDP is a control process that stochastically models decision-making scenarios; an agent continuously interacts with the environment by executing actions that change its internal state, and these actions are accordingly rewarded. The agent aims to improve its performance by progressively maximizing the received reward. Formally, a deterministic MDP is defined as $M=\langle S,A,R,s_0,T\rangle$,
where $S$ is a set of states, $A$ is a set of actions, $R : S \times A \rightarrow \mathbb{R}$ is a reward function that values how good or bad it is to take an action $a_t$ at a certain state $s_t$, $R(s_t,a_t) = r_t$, $s_0\in S$ is the initial state, and $T$ is a deterministic transition function $T: S \times A \rightarrow S$.

At each time step $t$, the agent takes an action $a_t \in A$ in state $s_t$ among all available actions following a \textit{policy} $\pi$ that maps states into a probability distribution over actions. In our work, $\pi(a|s)$ is an stationary stochastic policy $\pi: S \times A \rightarrow [0,1]$. For a state $s_t \in S$, the policy $\pi$ outputs an action $a_t$ with probability $\pi(a_t|s_t)$, which applied to $s_t$ returns $T(s_t,a_t)=s_{t+1}$ with reward $R(s_t,a_t)=r_t$. The objective is to learn an optimal policy $\pi^*$ that maximizes the expected cumulative discounted reward (formally shown in Equation (1)) where $s_0$ is the initial state and $\gamma \in [0,1]$ is the discount factor used to weight future rewards.

\begin{equation} 
    \pi^*=\argmax_{\pi} \mathbb{E}_\pi \left[\sum_{t\geq0}\gamma^t r_{t} \mid s_0\right]
\end{equation}

A Markov Game extends the concept of a Markov Decision Process to the multi-agent scenario \cite{markovgames}. It is defined as a tuple $G=\langle n,S,{A}_{1\dots n},R_{1\dots n},s_0,T\rangle$, with $n$ the number of agents, $S$ the set of states, $A_k$, $R_k$ the action set and reward function of agent $k$, respectively, $s_0\in S$ the initial state and $T$ the transition function. Policies in Markov Games are direct extensions of MDP policies; concretely, an optimal policy in a Markov Game is one that maximizes the collective reward of all agents. Markov Games in which all agents share the same reward function are defined as Multi-Agent MDPs (MMDPs), and its properties are the object of study of Centralized Multi-Agent Reinforcement Learning.

\begin{figure}[t]
    \centering
    \quad\quad\includegraphics[width=0.45\textwidth]{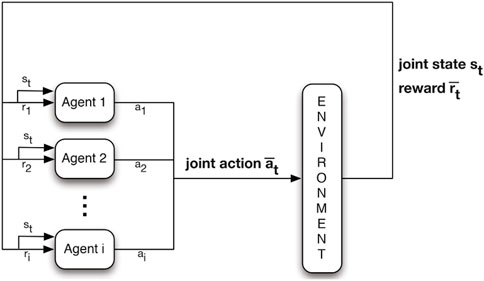}
    \caption{A centralized approach for the MARL problem \cite{jointmarl}.}
    \label{fig1}
\end{figure}

\begin{figure*}[t]
    \centering
    \includegraphics[width=\textwidth]{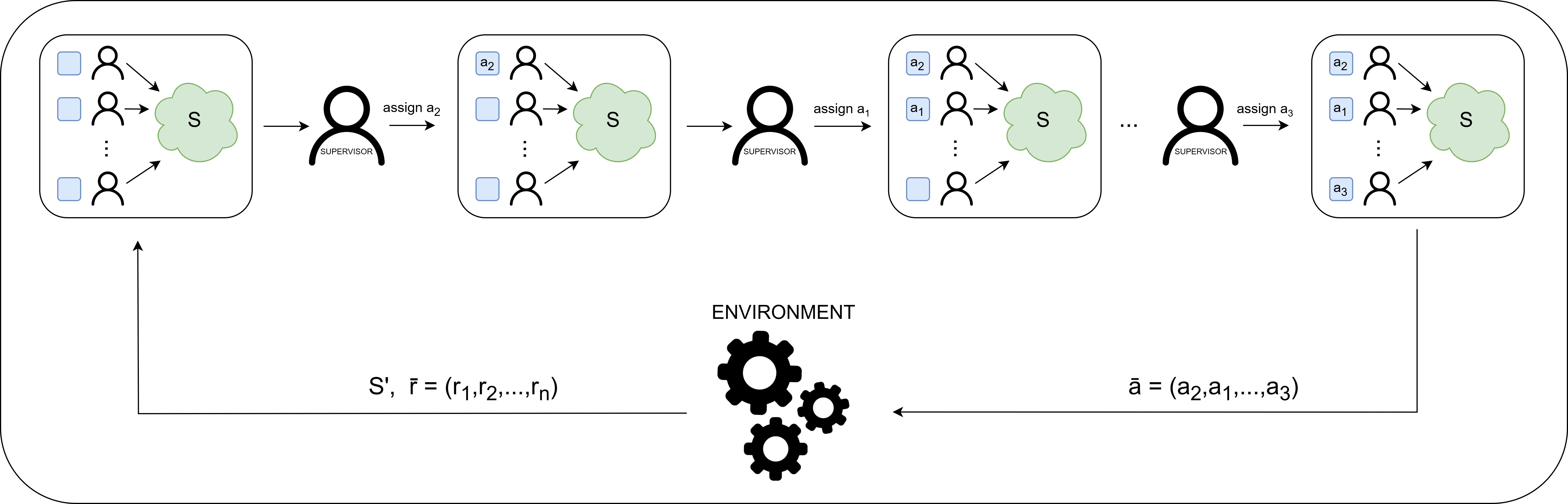}
    \caption{Example of a joint action construction using the supervisor agent.}
    \label{fig2}
\end{figure*}

\subsection{Centralized MARL}

In centralized approaches, the problem of finding an optimal policy for a Multi-Agent MDP can be reduced to a single MDP. Centralized approaches in MARL follow a scheme similar to the one shown in Figure \ref{fig1} \cite{jointmarl}. At time step $t$, a joint action $\overline{a}_t=(a_1,\dots,a_n)$, where $n$ is the number of agents in the environment and $a_i$ is the individual action that an agent $i$ can perform from a set of individual actions ${A}_i$, $a_i \in {A}_i$, is executed. Its execution produces a state $s_t$ that encapsulates the state variables of the agents and a reward $\overline{r}_t=(r_1,...,r_n)$, which comprises the individual reward of each agent $i$ after taking action $a_i$ in the environment.

The action space in centralized MARL thus becomes the cartesian product of every individual action ${A}_i$ over the number of agents, i.e.,  $A=\prod_{i=1}^n{A}_i$. In most of the MAL environments, agents usually share the same action space ${A}$ as the task of learning a coordination strategy typically requires agents with the same capabilities; for example, in a grid environment, each agent can move up, down, left and right. In this work, we adopt this assumption for readability purposes, and thus ${A}_i = {A}_j\ \forall i,j \in \{1,\dots,n\}$. 

Centralized approaches involve a virtual \textbf{learning entity} or single learner that acts as a communicator between the environment and the agents. Some approaches to centralized MARL include learning policy-factoring schemas \cite{cassano2021logical}; delegation of work and/or representation with coevolutionary approaches \cite{pottercoev97, pottercoev94}; reformulations of classical Reinforcement Learning algorithms such as DQN \cite{ong2015distributeddeepqlearning}; or hierarchical task allocation \cite{hierarchicalcooperative}.

\section{A supervising approach}
\label{supervision}

Dealing with the fully joint state-action space for all agents is, generally speaking, impractical for learning centralized policies in MARL. For example, in an environment with six agents, $Ag_1, Ag_2, Ag_3, Ag_4, Ag_5, Ag_6$, that can execute five different actions ${A} = \{a_1, a_2, a_3, a_4, a_5\}$,  the usual centralized approach with joint actions results in an action space of $|{A}|^6=5^6=15625$ actions. We can think of a joint action as an assignment of individual actions from the set ${A}$ to each agent in the environment; 
that is,  a joint action like $\overline{a}_t=(a_2,a_1,a_3,a_2,a_4,a_3)$ represents that agents $Ag_1$, $Ag_2$, $Ag_3$, $Ag_4$, $Ag_5$, $Ag_6$ are assigned actions $a_2,a_1,a_3,a_2,a_4,a_3$, respectively, which they will simultaneously execute at time $t$.
The number of joint actions grows exponentially with the number of agents, and shortly becomes intractable for the majority of methods including Deep Reinforcement Learning, which is known to struggle when dealing with large action spaces \cite{largediscrete}. 

Our proposal to alleviate this limitation, called \textbf{sequential construction of joint actions}, involves the definition of an {abstract agent} called \textit{supervisor}, responsible for sequentially assigning actions to the agents. Unlike the environment agents, the supervisor does not directly perform any action.

Following the previous example, the joint action $\overline{a}_t=(a_2,a_1,a_3,a_2,a_4,a_3)$  can be sequentially constructed by first assigning $a_2$ to $Ag_1$ ($assign\ a_2$), secondly assigning $a_1$ to $Ag_2$ ($assign\ a_1$), thirdly assigning $a_3$ to $Ag_3$ ($assign\ a_3$) and so on. Note that there is no need to specify which agent the action is assigned to, as the process of assigning actions to agents is done in agent order.

The supervisor's action set ${A}'$ is defined to be isomorphic to the individual action space ${A}$, and so it contains only $|{A}'|=|{A}|=5$ actions, namely, $assign\ a_1$, $assign\ a_2$, $assign\ a_3$, $assign\ a_4$, $assign\ a_5$, in the example. This change in the action space also requires a mechanism that allows the abstract agent to keep track of the actions that have been assigned, which introduces the need to enrich the original state space to reflect the assignments that have been made.

Our approach is illustrated in Figure \ref{fig2}; given an environment, the abstract agent learns to sequentially construct the joint action from the set actions ${A}'$ assigning one action to one agent at a time. To achieve this, the supervisor observes the current state (green cloud) together with the already assigned actions (light blue squares) at each time step. This observation, referred to as \textit{meta-state}, allows the supervisor to identify which agent remains unassigned and thus determines the behavior of the next to-be-assigned agent. Subsequent assignments are then reflected in the supervisor's already assigned actions (light blue squares with an action label in them), allowing it to tailor the behavior of each agent over time. This process implies that a single policy is learned for the supervisor. The supervisor's behavior will consist of the sequential construction of joint actions, implicitly incorporating sub-policies that specify the behavior of each agent within their respective action assignments.

As can be seen in the lower part of Figure \ref{fig2}, once all agents are assigned an action, the supervisor constructs the joint action and commands its execution in the environment, which produces the next state $S'$ and the next reward $\overline{r}$. The state $S'$ retrieved from the environment becomes then part of the subsequent meta-state, and the agents' action assignments are reset, starting again the whole process. Recall that in this abstraction, the supervisor does not need to have explicit knowledge of the agents' models (transition system), since its only purpose is to construct the joint action and command its execution.

\section{Implementation}
\label{implementation}

In this section, we will explore the implementation of the sequential construction of joint actions in a centralized MARL scenario. The process involves a compilation of the MMDP that models the original problem into a single-agent MDP that uses the notion of supervisor described in the previous section.

\subsection{Compilation}

Given an MMDP $M=\langle n,S,{A}_{1\dots n},R,s_0,T\rangle$ that models a multi-agent environment such that $A_i=A_j, \ \forall i,j \in \{1,\dots,n\}$, i.e., all agents have the same set of actions, which we will simply call $A$, we define a compilation process that transforms $M$ into an abstract MDP $M'=\langle S', A', R', s_0', T' \rangle$ using the  notion of supervisor introduced in section \ref{supervision}: 

\begin{itemize}
    \item States $s'\in S'$ are defined as meta-states $s'=(s,L)$, where $s \in S$ is a state from the original MMDP $M$ and $L=(o_i)_{i=1}^n$ is a sequence of actions to be applied in $s$, $o_i \in {A} \cup \{-\}$, where $[-]$ means that no action is assigned yet
    \item The set $A'$ of meta-actions is defined as  $A'=\{(assign\ a)\ |\ a \in A\}$
    \item The reward function $R'$, or meta-reward, is defined as follows: \begin{itemize}
        \item If $s'=(s,(a_{i_1},a_{i_2},\dots,a_{i_k},-,\dots,-))$, $s \in S$ and $a_{i_j} \in A$, that is, whenever the number of $[-]$ in the sequence is two or more, we define $R'(s',(assign\ a_j))=0$  as the assignment process is not finished yet.
        \item If $s'=(s,(a_{i_1},a_{i_2},\dots,a_{i_{n-1}},-))$, $s \in S$ and $a_{i_j} \in A$, that is, there is only one agent left to be assigned an action, we define $R'(s',(assign\ a_j))=\sum_{i=1}^n [R(s,(a_{i_1},\dots,a_{i_{n-1}},a_j))]_i$, i.e., the sum of all agents' rewards from the multi-agent environment.  
    \end{itemize}
    \item The initial state $s_0'$ is defined as the meta-state $s_0'=(s_0,L_0)$, where $L_0=(-)_{i=1}^n$
    \item The transition function $T'$ is defined as follows:
    \begin{itemize}
        \item If $s'=(s,(a_{i_1},a_{i_2},\dots,a_{i_k},-,\dots,-))$, $s \in S$ and $a_{i_j} \in A$, that is, whenever the number of $[-]$ in the sequence is two or more, we define $T'(s',(assign\ a_j))=(s,(a_{i_1},a_{i_2},\dots,a_{i_k},a_j,-,\dots,-))$, i.e., the supervisor assigns the action to the next agent in the sequence.
        \item If $s'=(s,(a_{i_1},a_{i_2},\dots,a_{i_{n-1}},-))$, $s \in S$ and $a_{i_j} \in A$, that is, there is only one agent left to be assigned an action, we define $T'(s',(assign\ a_j))=(T(s,(a_{i_1},\dots,a_{i_{n-1}},a_j)),(-)_{i=1}^n)$; i.e., the supervisor ends the action assignment and the joint action is commanded to be executed in the multi-agent environment.
        \end{itemize}
\end{itemize}

\floatname{algorithm}{Procedure}
\begin{algorithm}[t]
\caption{Description of a step of the sequential construction of a joint action, in which the supervisor applies a meta-action over a meta-state.}
\label{alg1}
\textbf{Input}: MARL environment \textit{e}, meta-state $(s_i,L_i)$, meta-action $(assign\ a)$  \\
\textbf{Output}: Next meta-state $(s_{i+1},L_{i+1})$, meta-reward $r$ 

\begin{algorithmic}[1]
\STATE \textbf{procedure} \textsc{step}$(e,(s_i,L_i),(assign\ a))$:
\begin{ALC@g} 
\IF{$L_i.countOcurrences(-) > 1$} 
    \STATE \textit{/* If there are still actions to be assigned, assign them to the tuple */}
        \STATE $L_{i+1} \gets L_i$
    \STATE $k \gets L_i.computeFirstPosition(-)$
    \STATE $L_{i+1}[k] \gets a $
    \STATE $s_{i+1} \gets s_i$
    \STATE $r \gets 0$
\ELSE 
    \STATE \textit{/* If next action is the last one, add it and compute next step using the original environment */}
        \STATE $k \gets L_i.computeLastPosition()$
    \STATE $L_i[k] \gets a$
    \STATE $s_{i+1},R \gets e.execute(s_i,L_i)$ 
    \STATE $r \gets R.aggregate()$
    \STATE $L_{i+1} \gets (-,\dots,-)$
\ENDIF
\STATE \textbf{return} $({s_{i+1}},L_{i+1}),r$
\end{ALC@g}
\end{algorithmic}
\end{algorithm}

Procedure \ref{alg1} is a pseudo-algorithm that shows a step of the sequential construction of a joint action in a MARL environment $e$. This operation amounts to one of the supervisor steps in the upper part of Figure \ref{fig2}. Given a meta-state and a meta-action, this procedure shows the resultant meta-state and meta-reward that the supervisor receives from applying the meta-action in the abstract single MDP defined in this section.

First, the algorithm checks whether the meta-action $(assign\ a)$ will finish the current assignment, by checking whether there are two or more actions to be assigned in the current meta-state $(s_i,L_i)$. If so, it means that the number of $[-]$ in the tuple $L_i$ is greater than one (line 2). Then, $L_i$ is copied into $L_{i+1}$ and we compute the first position, $k$, of $[-]$ in $L_i$, overriding $L_{i+1}[k]$ with the agent's action assignment $a$ (lines 4-6). With this, the current agent will already be assigned action $a$. The next state $s_{i+1}$ remains the same (line 7) and the meta-reward $r$ is set to zero (line 8), as the supervisor has not yet finished the assignment. 

Otherwise, if only one agent remains to be assigned an action (line 10), i.e., if there is only one $[-]$ in $L_i$, the last agent is assigned action $a$ (lines 11-12). Then, as every agent is already assigned an action, the joint action $L_i$ is commanded to be executed in state $s_i$ in the multi-agent environment $e$ (line 13). The execution yields the next state $s_{i+1}$ and the joint reward $R$ from the multi-agent environment, which is then aggregated to construct the meta-reward $r$ (line 14). Subsequently, the next tuple of to-be-applied actions $L_{i+1}$ is emptied (line 15) since the assignment process is reset. The meta-state $(s_{i+1},L_{i+1})$ and the meta-reward $r$ are then returned (line 17).

\begin{figure*}[t]
    \centering
    \begin{tabular}{ c c c}
               \includegraphics[align=c,width=0.225\textwidth]{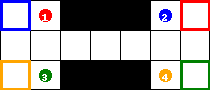} \quad \quad \quad \quad
        &       \includegraphics[align=c,width=0.225\textwidth]{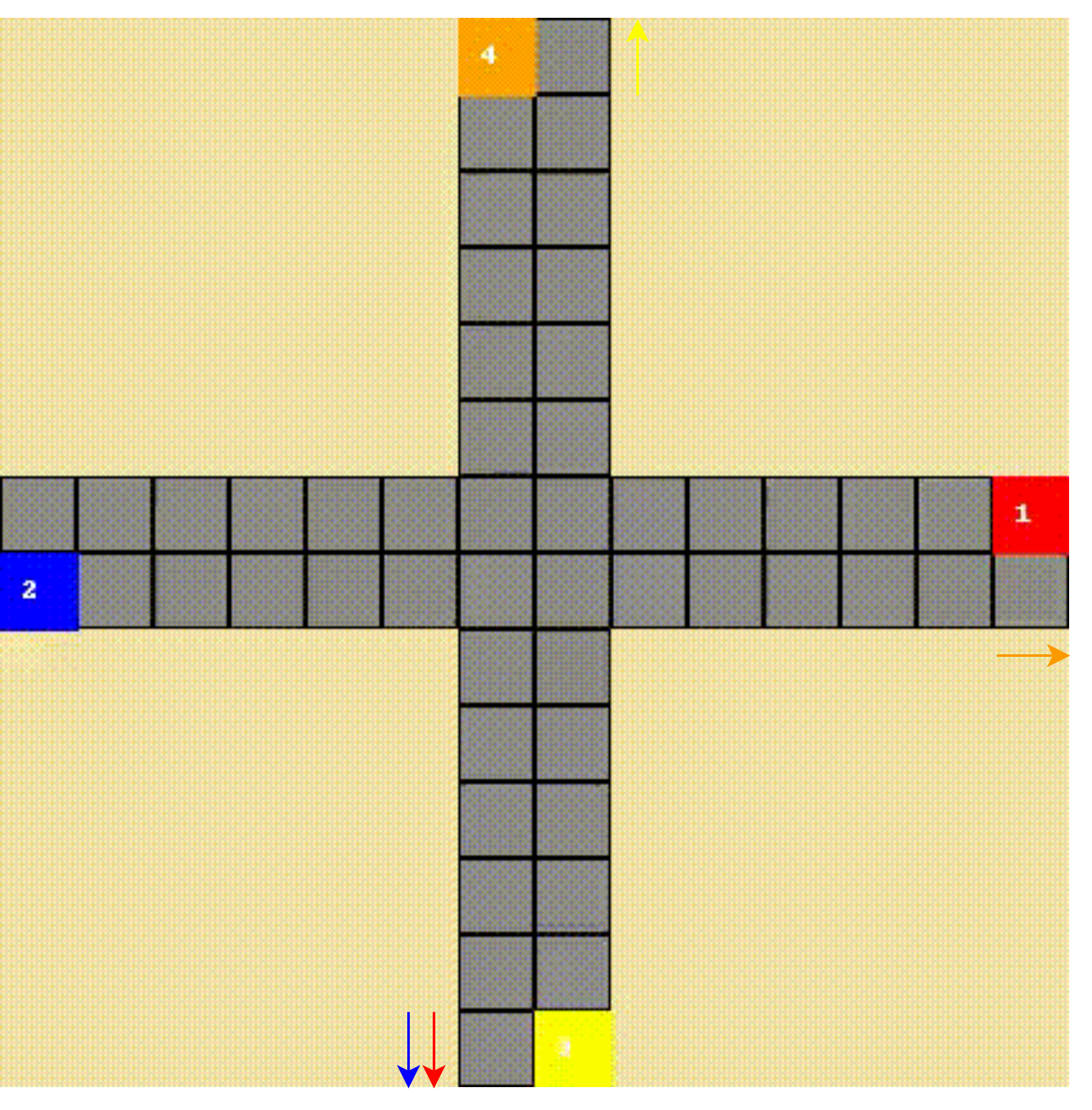} \quad \quad \quad \quad
        &       \includegraphics[align=c,width=0.225\textwidth]{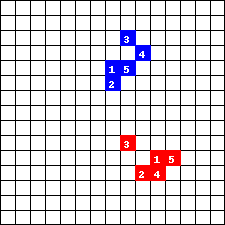}

    \end{tabular}
    \hfill 
    \hfill
    \\[\smallskipamount]
    \hfill
    \hfill
    \caption{An example of initial representation of the MARL environments. Respectively: a) \textbf{Switch}, b) \textbf{TrafficJunction} and c) \textbf{Combat}.}
    \label{fig3}
\end{figure*}

\subsection{Properties}

The main advantage of this approach is that it shifts the explosion of the joint action space to the state space. The joint actions now form part of the space of meta-states $S'$; as joint actions are no longer explicitly defined, they are now embedded in the meta-state definition. A state $s$ of the multi-agent environment is mapped to a set of meta-states, which includes the meta-state with no action assignment $(s,(-,\dots,-))$, the meta-states with one action assignment, for example, $(s,(a_2,-,\dots,-)$ or $(s,(a_3,-,\dots,-))$, the meta-states with two action assignments, for example, $(s,(a_2,a_1,-,\dots,-))$ or $(s,(a_1,a_1,-,\dots,-))$, and so on including all the meta-states composed of all possible combinations of actions to all the agents. For every partial assignment of $k$ agents there is a number of $|A'|^k$ meta-states, so a state is mapped to a total of $\sum_{i=0}^n|A'|^i$ meta-states. We can  further characterize the cardinality of the set $S'$ as $|S'|=|S|\cdot  \sum_{i=0}^n|A'|^i$.

At first glance, one might think that the explosion of the meta-state space of the supervisor abstraction would cause scalability problems similar to those induced by the joint action space of centralized approaches. However, the main difference is that Deep Reinforcement Learning methods have proven to be highly successful for complex continuous state spaces \cite{packer2019assessing}, but behave poorly with large action spaces \cite{largediscrete}. Numerous papers indicate that neural networks in large state spaces can establish certain correlations between states, thus facilitating the transfer of knowledge from one state to others that are similar \cite{witty2018measuring}. The generalization capabilities of DRL methods to complex environments proves that this approach is feasible, promising and, at least, worth analyzing.


\section{Experiments}
\label{experiments}

In this section, we present the experimental evaluation of the sequential construction of joint actions for several multi-agent environments using Reinforcement Learning. We define the abstract MDP explained in section \ref{implementation} for each MARL environment and 
analyze the overall performance of the supervisor agent over time. To conduct our experiments, we will use part of the collection of multi-agent environments available in \textsc{ma-gym} \cite{magym}. The multi-agent environments of \textsc{ma-gym} are based on \textsc{OpenAI gym}, an open-source library that provides a standard API to communicate between RL algorithms and environments as well as a standard set of environments for developing and comparing RL algorithms \cite{towers_gymnasium_2023}. \textsc{ma-gym} also allows to modify the agents' perception of the scenario. It allows observations to be defined as either a conjunction of local observations to the agents, which we will refer to as \textbf{individual} observations, or as a global observation to the whole scenario, which we will refer to as \textbf{collective} observations. For example, an agent can observe only its position in a grid or it can observe the grid as a whole.

\subsection{MARL Environments}

The \textsc{ma-gym} framework provides 8 configurable environments for MARL experimentation. We selected three environments that are particularly well-suited for cooperative tasks:   a) \textbf{Switch}, b) \textbf{TrafficJunction}, and c) \textbf{Combat} (see Figure \ref{fig3})

\textbf{Switch.} It is a grid-like environment with up to four agents. Each agent can move in any four directions (up, down, left, right) or stay still, and its target is to reach the tile of its same color, which is located at the opposite extreme of a corridor (Figure \ref{fig3}, environment a)). The agents must coordinate their moves to reach their targets as fast as possible and an agent cannot move to a tile occupied by another agent. When an agent reaches its destination, it is rewarded with +5. The cooperative task is for all agents to reach their destinations by coordinating their passage through the corridor.

\textbf{TrafficJunction.} It consists of a 4-way junction on a grid (Figure \ref{fig3}, environment b)). A car agent is said to enter the junction when it appears on the grid, and not every car enters the junction at the same time. A car entering the junction is assigned a route defined by an end position at the junction, represented by the colored arrows in Figure \ref{fig3}. The itinerary of each agent is fixed and depends on their route. A car can only do two things: move one cell forward on its route or brake and stay still. A collision is produced when two cars are in the same cell, which is negatively rewarded (-10). For each time step that an agent is in the junction, it receives a negative reward of ($-0.01\tau)$, where $\tau$ is the number of time steps elapsed since it entered the junction. The cooperative task is for every car to pass through the junction successfully without colliding. 

\textbf{Combat.} This environment is a simulation of a simple battle involving a blue team and a red team (Figure \ref{fig3}, environment c)). The blue team is the one that the environment controls, while the red team is automatically controlled and is external to the environment. The cooperative task is for the blue team to collectively defeat the red team, i.e., to defeat every red agent. A blue agent can move to another square in the four directions or attack a red agent (it will only hit if it is within 3 squares of the red agent). If an agent is attacked, it needs to take one-time recovery step before it can attack again. An agent also needs to be attacked three times to be defeated, so we will say that the blue and red agents each have three health points. There is a reward of -1 if the blue team loses at the end of the game, and an additional reward of -0.1 times the enemy team's total remaining health points. 

\begin{table}[t]
    \centering
    \caption{Tasks description for our experiments}
    \begin{tabular}{c c c c}
        Name & \# Agents ($n$) & Grid size & State space \\
        \hline
        \textbf{Switch$n$-v0} & [2-4] & (3,7) & individual \\
        \textbf{Switch$n$-v1} & [2-4] & (3,7) & collective \\
        \textbf{TrafficJunction$n$-v0} & \{4,7,10\} & (14,14) & individual \\
        \textbf{TrafficJunction$n$-v1} & \{4,7,10\} & (14,14) & collective \\
        \textbf{Combat$n$-v0} & [5-7] & (15,15) & individual \\
        \textbf{Combat$n$-v1} & [5-7] & (15,15) & collective \\
        
    \end{tabular}
    \label{tab1}
\end{table}

We created three tasks for each version of the environment varying the number of agents  (see Table \ref{tab1}). For the \textbf{Switch} environment, we created tasks \textbf{Switch2}, \textbf{Switch3} and \textbf{Switch4}, with 2, 3 and 4 agents, respectively. For the \textbf{TrafficJunction} environment, we created tasks \textbf{TrafficJunction4}, \textbf{TrafficJunction7} and \textbf{TrafficJunction10}, with 4, 7 and 10 agents, respectively. For the \textbf{Combat} environment, we created tasks \textbf{Combat5}, \textbf{Combat6} and \textbf{Combat7}, with 5, 6 and 7 agents, respectively. We also created two versions for each task, version \textbf{v0} using {individual}  observations, and version \textbf{v1} using {collective} observations. In total, there are 18 tasks.

\subsection{Training}

\begin{figure*}[t]
    \centering         \includegraphics[width=0.9\textwidth]{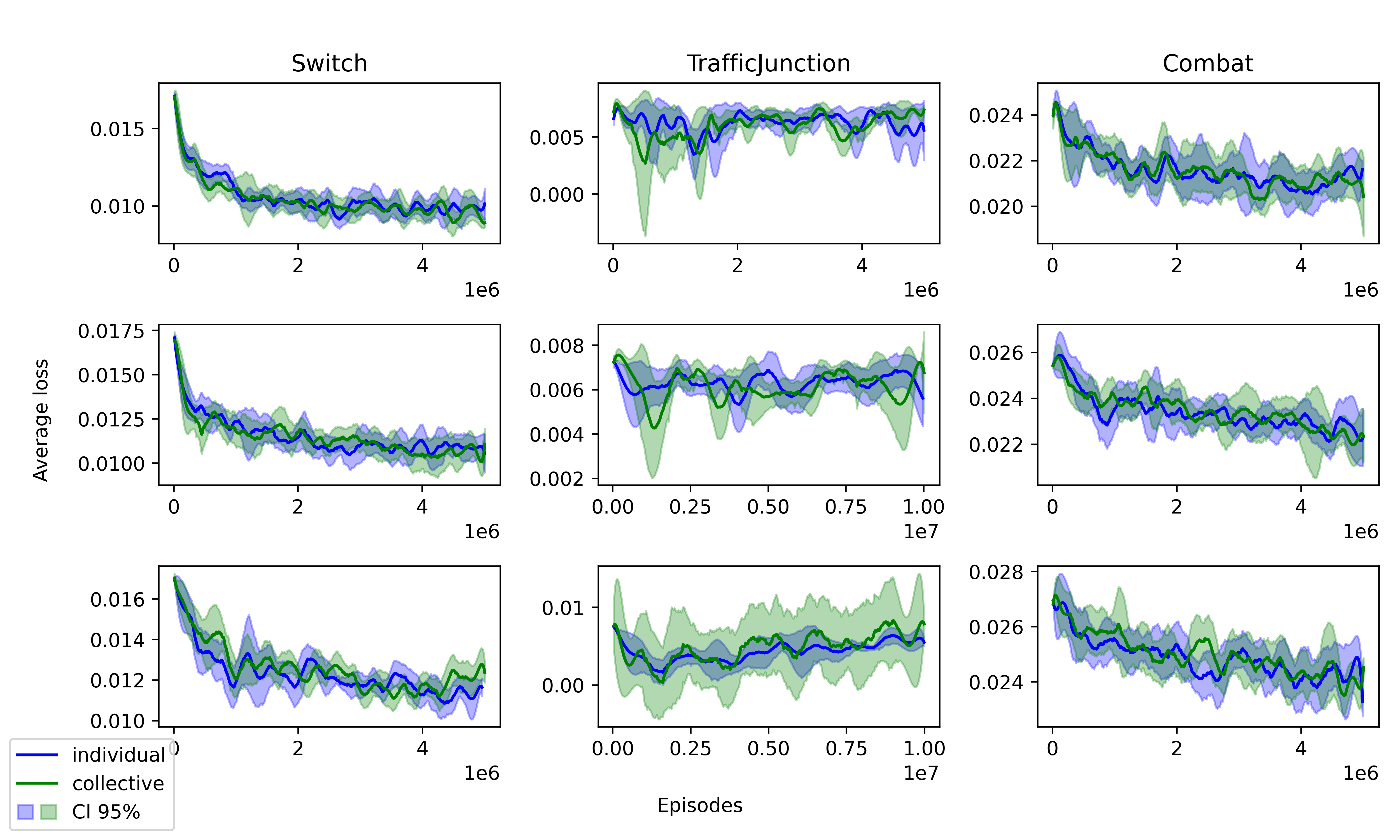}
         
    \caption{Evolution of average actor loss per episode for each environment and their three tasks}
    \label{fig4}
\end{figure*}

We trained a policy for the supervisor agent for each defined task, using Proximal Policy Optimization (PPO). PPO follows the philosophy of the actor-critic scheme \cite{actorcriticNIPS}, where the policy training involves an actor that explores the state space using a function approximator and a critic that evaluates the actor's performance using its beliefs, also using an approximator. The actor's approximator is trained using a special clipping function to prevent drastic changes from consecutive iterations. This cross-training actor-critic approach has become the state of the art in DRL \cite{baker2020emergent, duan2016benchmarking}.

The actor and critic networks are represented as feed-forward fully-connected neural networks, with 6 hidden units of sizes (input,output): $(M,256)$, $(256,256)$, $(256,128)$, $(128,128)$, $(128,64)$ and $(64,N)$, where $M$ is the input size and $N$ the output size. The actor network has as input the size of the observation. Following the implementation of procedure \ref{alg1}, $M=|s|+n$, where $n$ is the number of agents and $|s|$ is the size of a state of the multi-agent environment; the size of the output action space is $N=|A'|$. As for the critic network, $M=|s|+n$ and $N=1$. 

PPO was run up to 5,000,000 time steps (10,000,000 for \textbf{TrafficJunction7} and \textbf{TrafficJunction10}), 1,000 time steps per episode, 10,000 time steps per batch, 10 updates per iteration, a 0.2 epsilon value, a discount factor of 0.99, and an entropy regularization coefficient of 0.01. We used full batch updates and single advantage estimation. Calculations were refined using Pytorch Geometric library \cite{pytorch_geometric}. Both actor and critic networks were optimized using Adam, with learning rate $0,0002$. Experiments were conducted on a machine with a Nvidia GeForce RTX 3090 GPU, a 12th Gen Intel(R) Core(TM) i9-12900KF CPU and Ubuntu 22.04 LTS operating system.

For each task, we conducted five full PPO iterations, i.e., five models, in order to minimize the variance of the algorithm. Training statistics can be visualized in Figure \ref{fig4}. The first column corresponds to Switch tasks, starting with \textbf{Switch2} and ending with \textbf{Switch4}. The middle column corresponds to TrafficJunction tasks, starting with \textbf{TrafficJunction4} and ending with \textbf{TrafficJunction10}. The third column corresponds to Combat tasks, starting with \textbf{Combat2} and ending with \textbf{Combat4}.  Figure \ref{fig4} shows the evolution of the average loss over the total number of training episodes for each task, comparing the individual (blue) and collective (green) tasks within each graphic. The X axis represents the number of training episodes, with values scaled by $10^6$ (or $10^7$ in the case of the last two tasks of TrafficJunction). For example, a value of $2$ on the axis corresponds to $2 \cdot 10^6$ training episodes. The Y axis represents the average loss over the five runs of PPO, shown with a blue line for the individual tasks and a green line for the collective tasks, along with a $95\%$ confidence interval represented by a band of the same color. Each training took less than 6.5 hours on average to complete. 

Figure \ref{fig4} shows that the average loss is successfully minimized through the number of episodes in the Switch and Combat environments. Although the loss value for TrafficJunction is lower than for Switch or Combat, probably due to the lack of collisions in the early iterations, it is also very variable, even increasing at some points. This is probably due to the nature of the environment, where more rewarding situations imply greater stress at the junction, leading to potential collisions and situations with either very positive or very negative rewards very close to each other. Increasing the number of agents has little effect on the loss for Switch, except for a higher number of fluctuations. For Combat, the loss is slightly higher when the number of agents is increased, although the difference is negligible. For TrafficJunction, the loss increases slightly as the number of agents increases because there are more cars in the junction and therefore more collisions.
In all training experiments, including collective information in the state space seems irrelevant. Not only do the training results not improve on average and statistically overlap with the individual results, but in some cases, they actually worsen the individual perception results. For this reason, we will only focus on the individual perception tasks hereafter.

\subsection{Evaluation}

Once the supervisor models have been trained, we evaluate them on the same tasks they were trained for. 
We will evaluate the models based on the best and average reward for each task compared to the optimal reward. The optimal reward for {Switch} tasks is 5 (every agent reaches its destination), for {TrafficJunction} it is $-0.01\tau$ (zero collisions, where $\tau$ is the total number of time steps of the episode), and for {Combat} is 0 (blue team wins).

\begin{table}[t]
    \centering
    \caption{Evaluation results for each analyzed task}
    \begin{tabular}{c c c c c}
        Name & Best rew. & Avg. rew. & Best len. (JA) & Avg. len. \\
        \hline
        \textbf{Switch2} & 5 & 2.562 & 38 (19)& 97.62 \\
        \textbf{Switch3} & 5 & 2.073 & 75 (25) & 193.5 \\
        \textbf{Switch4} & 5 & -4.969 & 264 (66) & 384.8 \\
        \textbf{TrafficJunction4} & -0.34 & -1860.279 & 136 (34) & 295.2  \\
        \textbf{TrafficJunction7} & -0.75 & -1281.633 & 525 (75) & 920.46 \\
        \textbf{TrafficJunction10} & -0.92 & -2798.335 & 920 (92) & 1605.3 \\
        \textbf{Combat5} & 0 & -10.33 & 35 (7) & 179.25 \\
        \textbf{Combat6} & 0 & -12.65 & 60 (10) & 203.7 \\
        \textbf{Combat7} & 0 & -15.72 & 49 (7) & 205.66\\
        
    \end{tabular}
    \label{tab2}
\end{table}

Table \ref{tab2} shows the evaluation results. We ran 100 rollouts, 20 rollouts for each of the five models trained for a task, allowing up to 5,000 time steps per rollout. The first column of Table \ref{tab2} presents the best reward achieved in the 100 rollouts, while the second column presents the average reward across all 100 rollouts. The third column displays the total number of meta-actions in the rollout with the best reward, and the fourth column presents the average number of meta-actions for all rollouts. Additionally, the third column also shows in parentheses the number of joint actions resulting from the meta-action assignments; i.e. the number of meta-actions divided by the number of agents.

We can see in Table \ref{tab2} that the best reward for each task equals the optimal reward of its environment. Notably, the most complex task (Combat) is solved successfully and with a low number of meta-actions on average, whereas the medium-complexity task (TrafficJunction) shows very poor average reward results compared to the optimal values. The problem size does not seem to have a significant impact on any task, as the action space remains constant regardless of the number of agents.


\textbf{Discussion}. Our analysis reveals that the supervisor model consistently achieves excellent results across the three environments. However, the evaluation results for the TrafficJunction environment are poor.  Although the agents act independently in the three environments, in {TrafficJunction} the agents' actions are closely interrelated as the cars must coordinate to avoid collisions at the junction. A similar situation occurs in {Switch}, where the agents need to coordinate to pass the corridor without blocking each other. However, the low complexity of this task makes coordination less challenging. In {Combat}, however, we observe the opposite: the agents can act freely because they play as a team, but the actions of the agents are independent of each other.

We believe that a high level of interaction between agents may be hindering the model's performance in domains such as TrafficJunction. We hypothesize that our approach is biased toward agents acting independently of one another since better results are obtained for Combat even being the most complex scenario. Overall, we believe that our model successfully fulfills its purpose and lays the groundwork for future research.

\section{Conclusions}
\label{conclusion}

We introduced a novel approach called \textbf{sequential construction of joint actions} in the context of MAL to address the challenge of coordinating multiple agents by consolidating individual agent actions into joint actions using an abstract supervisor agent. Abstracting the execution of joint actions into a sequential action assignment through an MDP compilation alleviates the explosion of the action space of centralized MARL tasks. The results of the experimental evaluation show that the supervising agent effectively learns to coordinate actions among agents, leading to stable and competitive performance across different task configurations. These results highlight the potential of our framework to improve the scalability and efficiency of MAL systems. Overall, our work not only presents a useful approach to the coordination problem in MARL but also provides empirical evidence of its effectiveness in diverse and challenging environments with different characteristics. 

As for \textbf{future work}, we will explore tasks with more interactions between agents, investigate other techniques to efficiently reduce the action space, and solve other types of problems. We would also like to compare our approach extensively with those based on the joint action space to determine the advantages and limitations of our approach.

\section{Acknowledgments}

This work is supported by the Spanish AEI PID2021-127647NB-C22 project and Ángel Aso-Mollar is partially supported by the FPU21/04273.

\bibliographystyle{IEEEtran}
\bibliography{biblio}

\end{document}